\documentclass{article}

\usepackage[final,nonatbib]{nips_2017}

\usepackage[utf8]{inputenc} 
\usepackage[T1]{fontenc}    
\usepackage{hyperref}       
\usepackage{url}            
\usepackage{booktabs}       
\usepackage{amsfonts}       
\usepackage{nicefrac}       
\usepackage{microtype}      
\usepackage{graphicx}
\usepackage{wrapfig}
\usepackage[dvipsnames]{xcolor}
\usepackage{subfig}

\newcommand{\hl}{\mathbf{h}}
\newcommand{\ul}{\mathbf{u}}
\newcommand{\rl}{\mathbf{r}}

\newcommand{\xl}{\mathbf{x}}
\newcommand{\zl}{\mathbf{z}}

\title{MinimalRNN: Toward More Interpretable and Trainable Recurrent Neural Networks}

\author{
  Minmin Chen \\
 Google\\
 Mountain view, CA 94043 \\
  \texttt{minminc@google.com} \\
}

\begin{document}

\maketitle

\begin{abstract}
We introduce MinimalRNN, a new recurrent neural network architecture that achieves comparable performance as the popular gated RNNs with a simplified structure. It employs minimal updates within RNN, which not only leads to efficient learning and testing but more importantly better interpretability and trainability. We demonstrate that by endorsing the more restrictive update rule, MinimalRNN learns disentangled RNN states.  We further examine the learning dynamics of different RNN structures using input-output Jacobians, and show that MinimalRNN is able to capture longer range dependencies than existing RNN architectures.  
\end{abstract}

\section{Introduction}
\label{sec:intro}

Recurrent neural networks have been widely applied in modeling sequence data in various domains, such as language modeling~\cite{mikolov2010recurrent,kiros2015skip,jozefowicz2016exploring}, translation~\cite{bahdanau2014neural}, speech recognition~\cite{graves2013speech} and recommendation systems~\cite{hidasi2015session,wu2017recurrent}. Among them, Long Short-Term Memory networks (LSTM)~\cite{hochreiter1997long} and Gated Recurrent Units (GRU)~\cite{chung2014empirical} are the most prominent model architectures. Despite their impressive performance, the intertwine and recurrent nature of update rules  used by these networks has prevented us from gaining thorough understanding of their strengths and limitations~\cite{karpathy2015visualizing}. 

Recent work on Chaos Free Networks (CFN)~\cite{laurent2017chaos} inspected these popular networks from a dynamical system viewpoint and  pointed out that existing RNNs,  including vanilla RNNs, LSTM and GRUs, intrinsically embody irregular and unpredictable dynamics. Even without interference from input (external) data, the forward trajectories of states in these networks attract to very different points with a small perturbation of the initial state. Take GRUs as an example, it updates its states over time as follows:
\begin{eqnarray}
\mathbf{h}_t = \mathbf{u}_t \odot \mathbf{h}_{t-1} + (\mathbf{1} - \mathbf{u}_t) \odot \tanh(\mathbf{W}_h (\mathbf{r}_t \odot \mathbf{h}_{t-1}) + \mathbf{W}_x \mathbf{x}_t + \mathbf{b}_h) 
\end{eqnarray}
where $\mathbf{u}_t$ and $\mathbf{r}_t$  are the update and reset gates respectively. 
The authors identified the multiplication $\mathbf{W}_h (\mathbf{r}_t \odot \mathbf{h}_{t-1})$ in the second part of the update, i.e., mixing of different dimensions in the hidden state, as the cause of the chaotic behavior. To address that, the authors proposed CFN, which updates its hidden states as 
\begin{eqnarray}
\mathbf{h}_t = \mathbf{u}_t \odot \tanh(\mathbf{h}_{t-1}) + \mathbf{i}_t \odot \tanh(\mathbf{W}_x \mathbf{x}_t + \mathbf{b}_x).
\end{eqnarray}
Here $\mathbf{u}_t$ and $\mathbf{i}_t$ are the update and input gates. By ruling out the mixing effect between the different dimensions in the hidden state, the network presents a much more predictable dynamic. The simpler network achieves comparable performance as the more dynamically complex LSTMs or GRUs for various  language modeling tasks. 

Inspired by the success of CFN, we propose another recurrent neural network architecture named Minimal Recurrent Neural Networks (MinimalRNN), which adopts minimum number of operations within RNN without sacrificing performance. Simplicity not only brings efficiency, but also interpretability and trainability.  
There have been evidences that favorable learning dynamic in deep feed-forward networks arises from input-output Jacobians whose singular values are $O(1)$~\cite{saxe2013exact,pennington2017resurrecting}. We empirically study the input-output Jacobians in the scope of recurrent neural networks and show that MinimalRNN is more trainable than existing models. It is able to propagate information back to steps further back in history, that is, capturing longer term dependency. 


\section{Method}
\label{sec:method}

\begin{wrapfigure}{r}{0.45\textwidth}
\vspace{-0.2in}
  \begin{center}
    \includegraphics[width=0.42\textwidth]{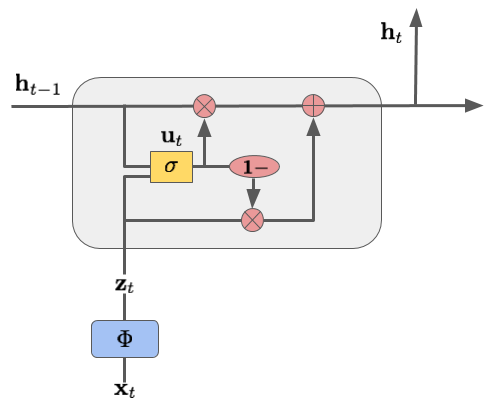}
  \end{center}
  \caption{Model architecture of MinimalRNN.}
  \label{fig:mrnn}
\end{wrapfigure}

Figure~\ref{fig:mrnn} illustrates the new model architecture named MinimalRNN. It trades the complexity of the recurrent neural network with having a small network outside to embed the inputs and take minimal operations within the recurrent part of the model. 

At each step $t$, the model first maps its input $\mathbf{x}_t$ to a latent space through
$$\mathbf{z}_t = \Phi(\mathbf{x}_t)$$
$\Phi(\cdot)$ here can be any highly flexible functions such as neural networks. In our experiment, we take $\Phi(\cdot)$ as a fully connected layer with $\tanh$ activation.  That is, $ \Phi(\mathbf{x}_t) = \tanh(\mathbf{W}_x\mathbf{x}_t + \mathbf{b}_z)$.

Given the latent representation $\mathbf{z}_t$ of the input, MinimalRNN then updates its states simply as:
\begin{eqnarray}
\mathbf{h}_t = \mathbf{u}_t \odot \mathbf{h}_{t-1} + (\mathbf{1}-\mathbf{u}_t) \odot \mathbf{z}_t 
\end{eqnarray}
where $\mathbf{u}_t = \sigma(\mathbf{U}_h \mathbf{h}_{t-1} + \mathbf{U}_z \mathbf{z}_t + \mathbf{b}_u) $ is the update gate. 

\textbf{Latent representation.} The dynamic of MinimalRNN prescribed by these updates is fairly straight-forward. First, the encoder  $\Phi(\cdot)$ defines the latent space. The recurrent part of the model is then confined to move within this latent space.  At each step $t$, MinimalRNN takes its previous state $\mathbf{h}_{t-1}$ and the encoded input $\mathbf{z}_t$, then simply outputs a weighted average of both depending on the gate $\mathbf{u}_t$.  That is,  dimension $i$ of the RNN state $h_t^i$  is activated by input $z^i_t$ and relax toward zero without any new input from that dimension. The rate of relaxation is determined by the gate $u_t^i$. It gets reactivated once it sees $z^i_t$ again.

Comparing with LSTM, GRU or CFN, MinimalRNN resorts to a much simpler update inside the recurrent neural network. It retains the gating mechanism, which is known to be critical to preserve long range dependencies in RNNs. However, only one gate remains. The update rule bears some similarity to that of CFN, in that both forbid the mixing between different dimensions of the state. 

\textbf{Trainability.} Recurrent neural networks are notoriously hard to train due to gradient explosion and vanishing~\cite{pascanu2013difficulty,collins2016capacity}.  Several recent works~\cite{saxe2013exact,xie2017all,pennington2017resurrecting} study information propagation in deep networks and suggest that well-conditioned input-output Jacobians leads to desirable learning dynamic in deep neural networks. In particular, if every singular value of the input-output Jacobians remains close to 1 during learning, then any error vector will preserve its norm back-propagating through the network. As a result, the gradient will neither explode nor vanishing. Thanks to the simple update rule employed in MinimalRNN, we can easily write out the input-output Jacobian, i.e., derivatives of RNN state $\hl_t$ w.r.t. input $\xl_{t-k}$ as follows: 
\begin{eqnarray}
\frac{\partial \hl_t}{\partial \xl_{t-k}}  & = & \left(\prod_{t-k < i \le t} \frac{\partial \hl_i}{\partial \hl_{i-1}}\right)\frac{\partial \hl_{t-k}}{\partial \zl_{t-k}} \frac{\partial \zl_{t-k}}{\partial \xl_{t-k}}\\
\mbox{where}\quad\frac{\partial \hl_i}{\partial \hl_{i-1}} & =  &  D_{\ul_i} + D_{(\hl_{i-1} - \zl_i)\odot \ul_i \odot (\mathbf{1} - \ul_i)}\mathbf{U}_h \nonumber
\end{eqnarray}
Here $D_{\ul}$ denotes a diagonal matrix with vector $\ul$ as the diagonal entries. Assuming the weight matrix $\mathbf{U}_h$ is unitary, that is, the singular values of $\mathbf{U}_h$ are all $1$, then we can easily see that the maximum singular value of $\frac{\partial \hl_i}{\partial \hl_{i-1}}$ is bounded by 1. Similarly for $\frac{\partial \hl_{t-k}}{\partial \zl_{t-k}}$. 

In comparison, the Jacobian of GRU is much more complex, 
\begin{equation}\label{eq:jacob_gru}
\frac{\partial \hl_i}{\partial \hl_{i-1}}  =    D_{\ul_i} + D_{(\hl_{i-1} - \zl_i')\odot \ul_i \odot (\mathbf{1} - \ul_i)}\mathbf{U}_h + D_{(\mathbf{1}-\ul_i)\odot(1-\zl_i'^2)}\mathbf{W}_h (D_{\rl_i} + D_{\mathbf{h}_{i-1}\rl_i(1-\rl_i)}\mathbf{R}_h)
\end{equation}
here $\zl_i' = \tanh(\mathbf{W}_h (\mathbf{r}_t \odot \mathbf{h}_{i-1}) + \mathbf{W}_x \mathbf{x}_i + \mathbf{b}_h)$ and $\mathbf{R}_h$ is the weight matrix used in the reset gate of GRU. The Jacobian has the additional multiplication term between $\mathbf{W}_h$ and $\mathbf{R}_h$ in each layer, which we hypothesize will result in GRUs more prone to exploding or vanishing gradient.

\section{Experiments}
\label{exp}

We demonstrate the efficacy of our method on a recommendation task of production scale. The goal is to recommend to users items of interest given user's historical interactions with items in the system.  

\textbf{Dataset.} The dataset contains hundreds of millions of user records. Each one is a sequence of (itemId, pageId, time) tuples, recording the context under which a recommended item consumed by an user. We consider user activities up to several months and truncate the sequence to a maximum length of 500.  The item vocabulary contains 5 million most popular items of the last 48 hours.  

\begin{figure}
  \begin{center}
    \includegraphics[width=0.32\textwidth]{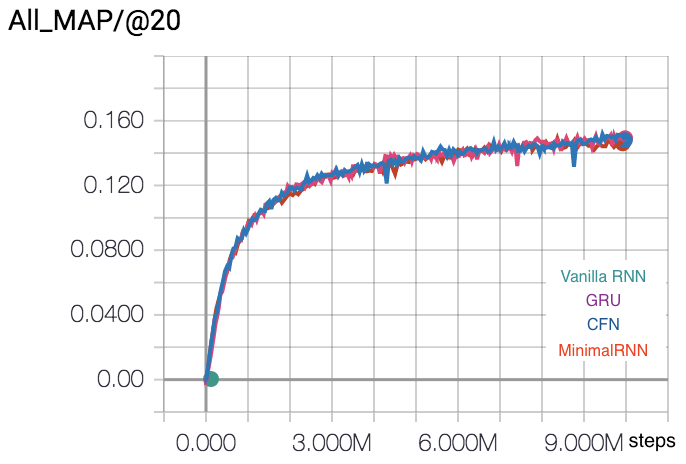}
   \includegraphics[width=0.32\textwidth]{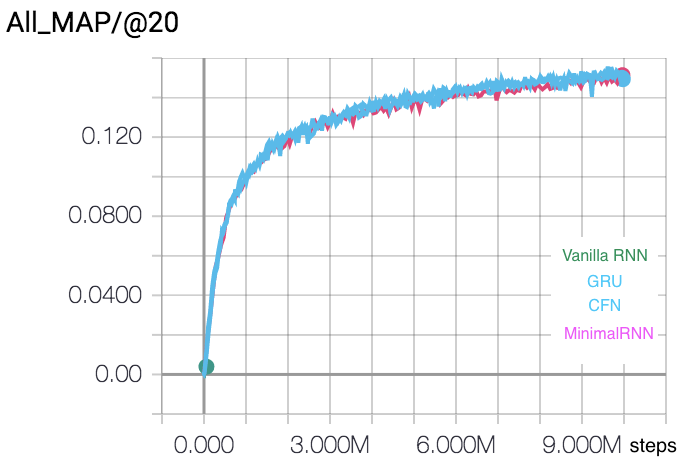}
   \includegraphics[width=0.32\textwidth]{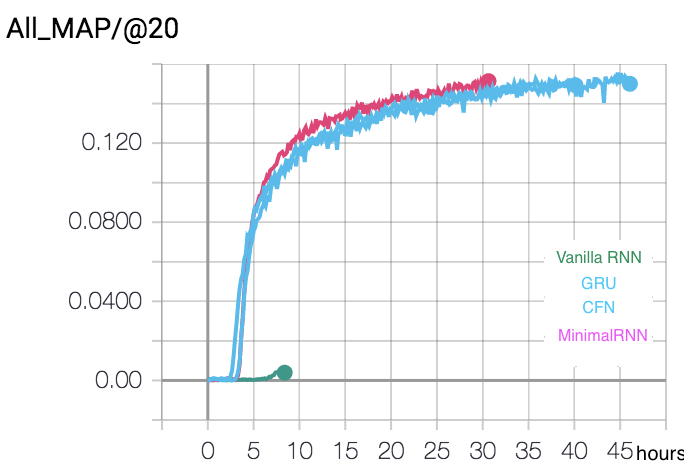}
  \end{center}
  \caption{MAP@20 evaluated on the test sets progressed over 10M learning steps.}
  \label{fig:map20}
\end{figure}

\begin{wrapfigure}{r}{0.45\textwidth}
\vspace{-0.3in}
  \begin{center}
    \includegraphics[width=0.44\textwidth]{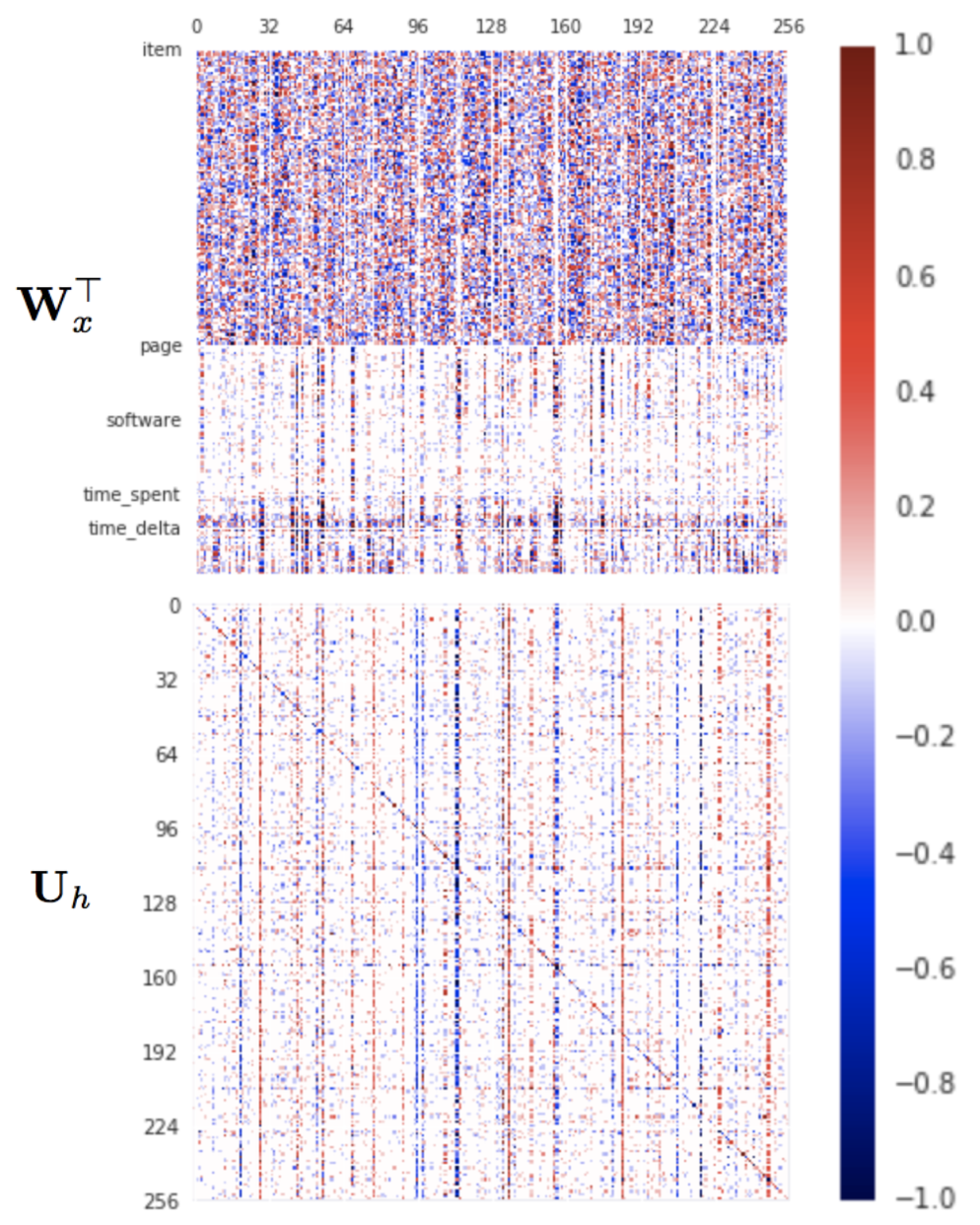}
  \end{center}
  \caption{(Top). Weight matrix $\mathbf{W}_x$ that transform input to latent space; (Bottom). Weight matrix $\mathbf{U}_h$ that computes the update gate according to previous hidden state $\hl_{t-1}$.}
  \label{fig:latent}
\end{wrapfigure}

\textbf{Setup.} Our production system uses  a GRU as the core mechanism to capture the evolving of user interest through time. The model takes a sequence of user actions on items, and aims to predict the next item the user is going to consume.  We replace the GRU component with various recurrent neural networks architectures, such as Vanilla RNN, CFN and MinimalRNN, and compare their performance to the production system. The main performance metric that we monitor offline is the Mean-Average-Precision@20.

\textbf{Performance.} Figure~\ref{fig:map20} plots the MAP@20 of the recommender system with different recurrent neural networks over 10M learning steps. All the weights in the recurrent neural nets are initialized to be unitary. As our data is refreshed daily, we were able to compare these methods over multiple datasets.  Figure~\ref{fig:map20} left and middle show two runs. In both cases, Vanilla RNN failed during early stage of the learning due to gradient explosion.  The other three models perform rather similar, reaching MAP@20 of 0.15. Since the update in MinimalRNN is much simpler, it takes less time to train. Learning finished in 30 hours comparing with CFN which takes 36 hours and 46 hours for GRU, as shown in the right column of figure~\ref{fig:map20}. 

\textbf{Latent representation.} In this experiment, we attempt to look inside the hidden states learned by MinimalRNNs. Our intuition is that the stricter updates in MinimalRNN forces its states to reside in some latent space defined by the input encoder $\Phi(\cdot)$. Each dimension of the state space focuses on some factor of the input, but not the others. 

The first row of figure~\ref{fig:latent} plots the weight matrix $\mathbf{W}_x^\top$ that is used to transform the input to the latent space. Each row is one input dimension and each column is one dimension in the latent space. Entry $(i, j)$ indicates the activation of input feature $i$ on the latent dimension $j$. The input are grouped by blocks, the first block is the item embedding, and second block is the page embedding, etc.. We can see that most of the latent dimensions are used to capture the item embedding, while remaining ones capture the other contexts, such as the page on which the item is displayed, the software the user used to access the item, and time information. The bottom row  of figure~\ref{fig:latent} plots the weight matrix $\mathbf{U}_h$ that is used to compute the update gate. Each entry $(i, j)$ indicates the activation of previous state $h_{t-1}^j$ on the forget gate entry $u_t^i$.  It shows several interesting properties. First, we observe strong activations on the diagonal. That is, the rate of forgetting depends mostly on the previous state from the same dimension, $h_{t-1}^i \rightarrow u_t^i$. Second, we can observe several dimensions (columns) having strong activations across all rows. In other words, these dimensions impact the rate of forgetting for almost all the dimensions in the hidden states, and these dimensions mostly corresponds to the dimensions that are capturing the context information as shown in the top of the figure. 

\begin{figure}
Vanilla\quad \includegraphics[width=0.22\textwidth]{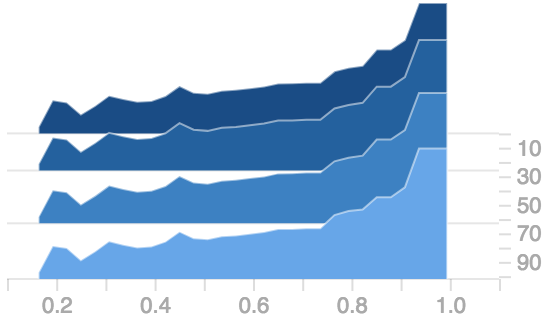} \includegraphics[width=0.22\textwidth]{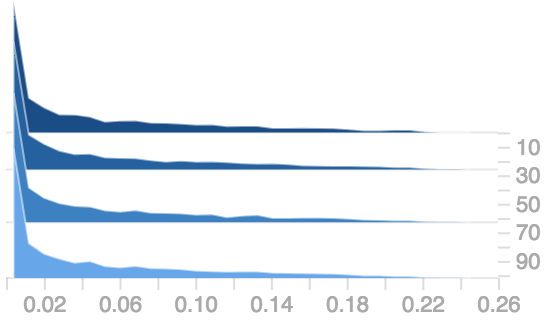}   \includegraphics[width=0.22\textwidth]{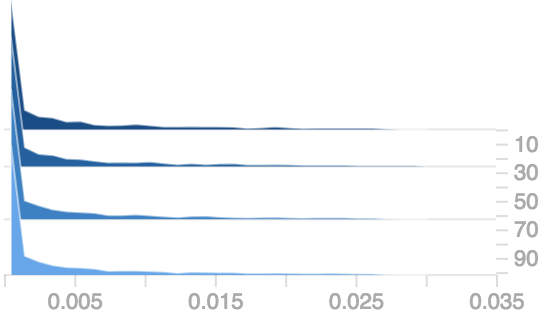} \includegraphics[width=0.22\textwidth]{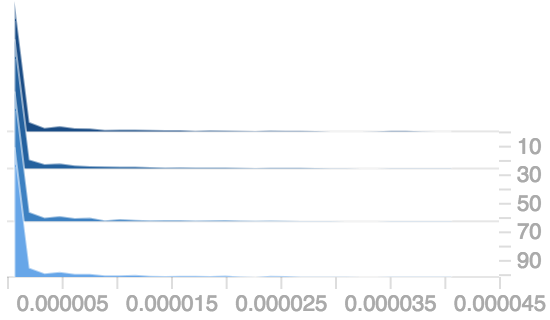} \\
GRU\qquad \includegraphics[width=0.22\textwidth]{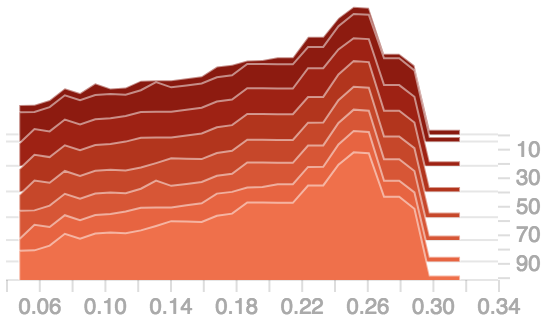}   \includegraphics[width=0.22\textwidth]{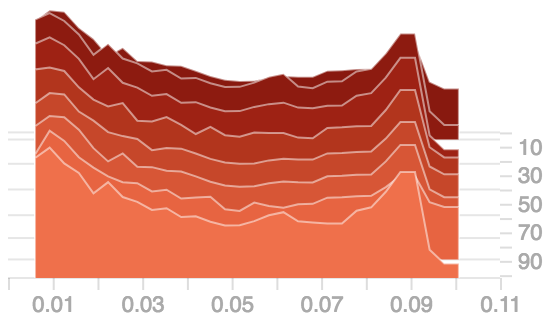}   \includegraphics[width=0.22\textwidth]{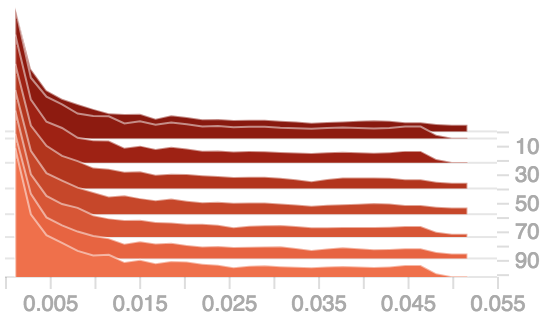}  \includegraphics[width=0.22\textwidth]{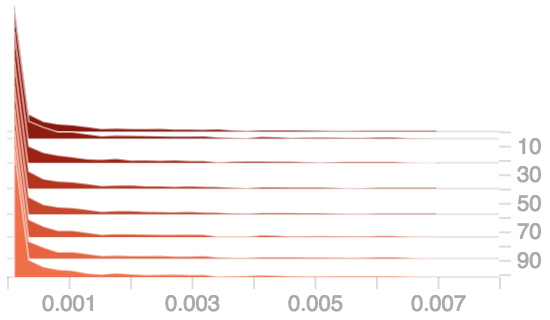}\\
CFN\qquad \includegraphics[width=0.22\textwidth]{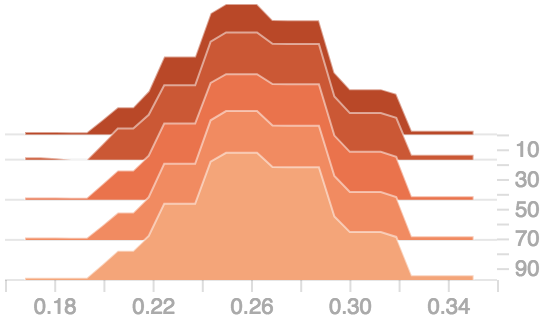}  \includegraphics[width=0.22\textwidth]{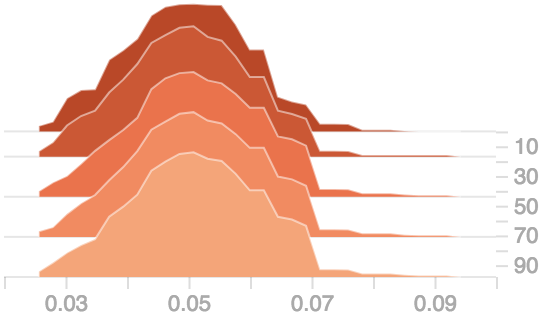}  \includegraphics[width=0.22\textwidth]{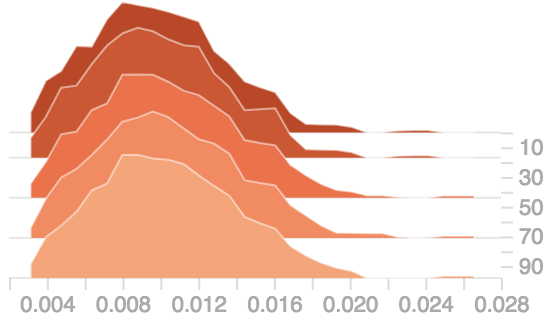}  \includegraphics[width=0.22\textwidth]{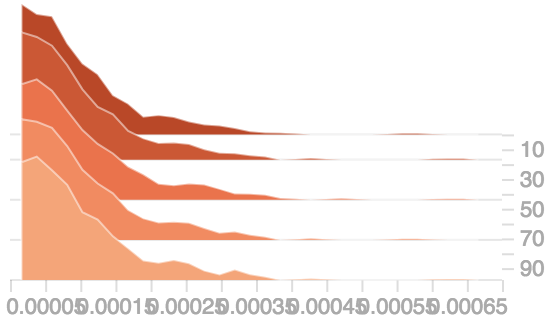}\\
Minimal  \subfloat[k=0]{\includegraphics[width=0.22\textwidth]{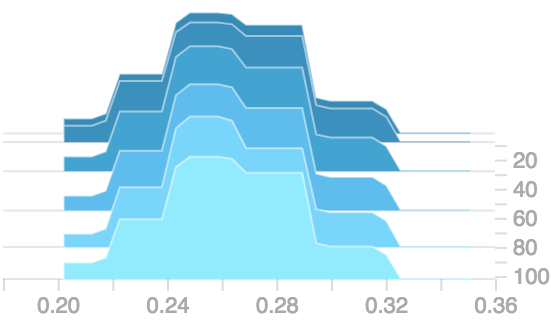}}  \subfloat[k=5]{ \includegraphics[width=0.22\textwidth]{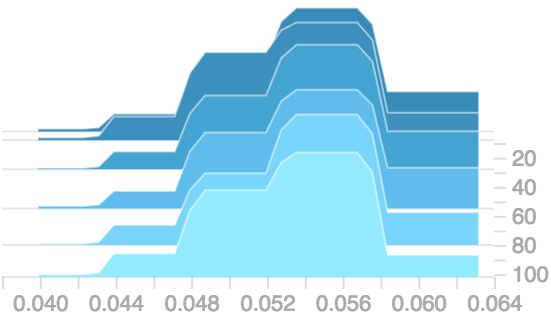}}   \subfloat[k=10]{\includegraphics[width=0.22\textwidth]{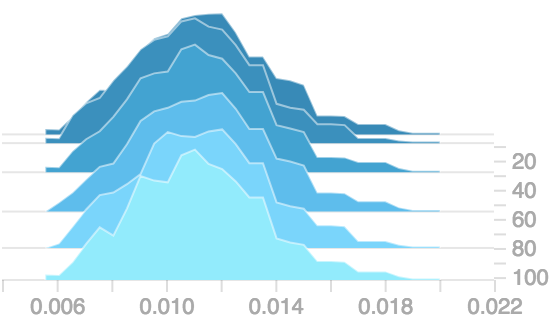}}   \subfloat[k=25]{\includegraphics[width=0.22\textwidth]{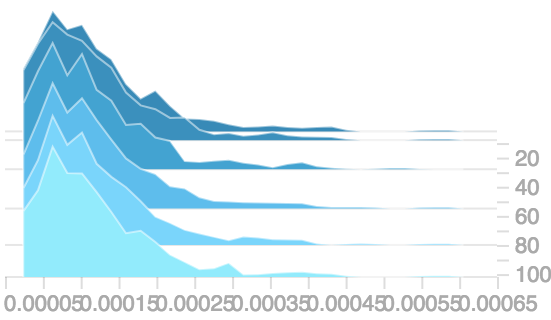}}
\caption{Histograms of the singular values of input-output Jacobians $\frac{\partial \hl_T}{\partial \xl_{T-k}}$ for $k = 0, 5, 10, 25$ at initial point, with weight matrices random initialized to be unitary.}
\label{fig:svd_init}
\end{figure}

\textbf{Trainability.} In these experiments, we take the input-output Jacobians computed from different recurrent neural networks at initial point and during learning to understand the trainability of these RNN structures. Figure~\ref{fig:svd_init} plots the histogram of the singular values of the Jacobian matrix over various $k$ at \textbf{initial point}. All the weights in the networks are initialized to be unitary. When $k=0$, we are looking at the derivatives of the RNN hidden state w.r.t. the current input, while $k=25$ depicts the derivatives w.r.t. input that is 25 step back. We can see that  the singular values of the input-output Jacobians in vanilla RNN quickly vanishes towards zero as $k$ increases. The singular values of the input-output Jacobians for the GRUs starts to stretch in some directions, and shrink in others when $k$ reaches 10, which we hypothesize is due to the additional multiplication as shown in equation~(\ref{eq:jacob_gru}). The input-output Jacobians of CFN and MinimalRNN on the other hand are relatively well-conditioned even for $k = 25$. Especially for MinimalRNN, the singular values stay normally distributed as $k$ increases, and neither stretching nor shrinking in any directions. 

\begin{figure}
\vspace{-0.2in}
  \begin{center}
  \includegraphics[width=0.32\textwidth]{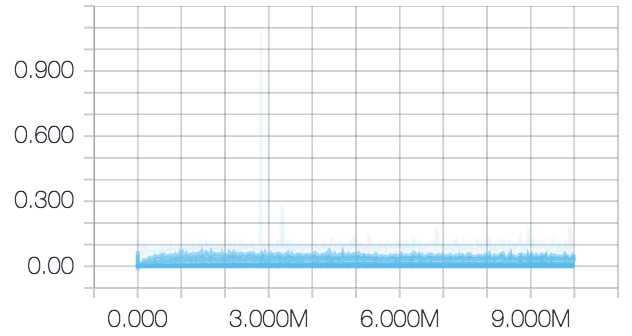}
  \includegraphics[width=0.32\textwidth]{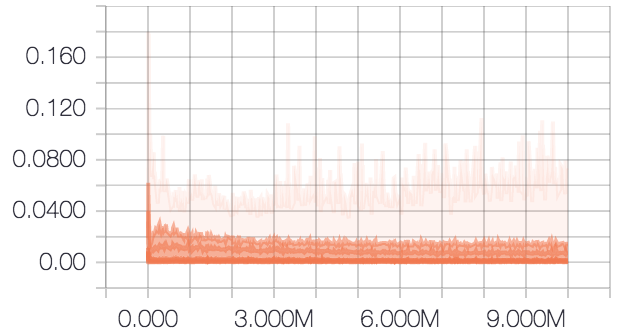}
  \includegraphics[width=0.32\textwidth]{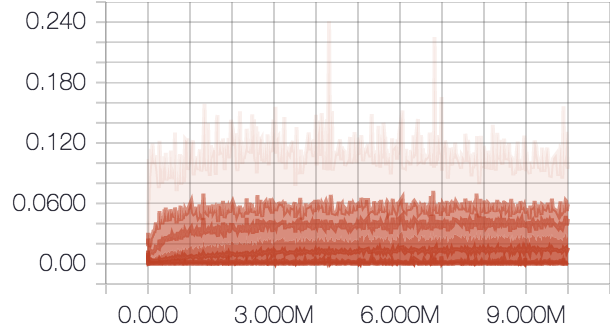}\\
   \subfloat[GRU]{\includegraphics[width=0.32\textwidth]{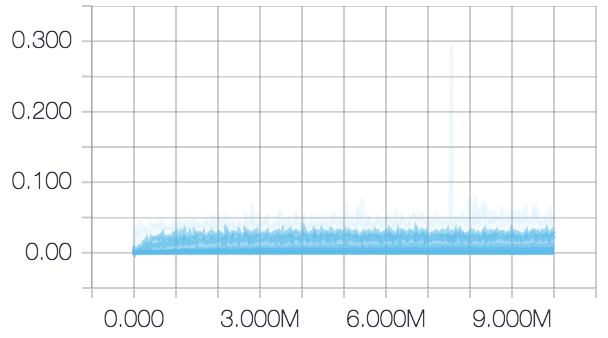}}
    \subfloat[CFN]{\includegraphics[width=0.32\textwidth]{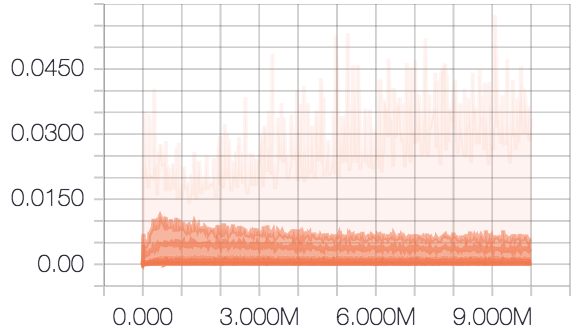}}
    \subfloat[MinimalRNN]{\includegraphics[width=0.32\textwidth]{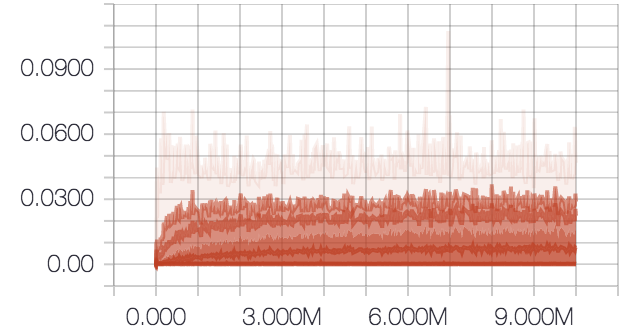}}
  \end{center}
  \caption{Distribution of singular values of input-output Jacobian $\frac{\partial \hl_T}{\partial \xl_{T-k}}$ at $k=10$ (first row) and $k=25$ (second row) during 10M learning steps. Each line on the chart represents a percentile in the distribution over the data: for example, the top line shows how the maximum value has changed over time, and the line in the middle shows how the median has changed. Reading from top to bottom, the lines have the following meaning: [maximum, 93\%, 84\%, 69\%, 50\%, 31\%, 16\%, 7\%, minimum]. }
  \label{fig:svd_learn}
  \vspace{0.1in}
\end{figure}

As pointed out in~\cite{xie2017all}, a good initialization does not necessarily guarantee trainability. Figure~\ref{fig:svd_learn} plots the distribution of the singular values of the Jacobian matrices throughout the whole learning process. As learning of Vanilla RNN failed quite early, we ignore it in the comparison. We can see that the singular values of the Jacobian matrix in GRU grows rapidly in some iterations, suggesting the back-propagation error could be stretching over those directions. The singular values of the Jacobians in CFN are shrinking mostly toward 0 as learning goes on. In comparison, the Jacobians of MinimalRNN are relatively well-conditioned throughout the learning. We can observe similar trends for different values of $k, k > 10$. These results suggest that MinimalRNN could be able to capture input far back in the history, i.e., longer range dependencies.

\section{Future work} 

It remains to be seen if this extremely simple recurrent neural network architecture is able to carry over to a wide range of tasks besides the one presented here. Our most performant model for this task so far only uses one fully connected layers in $\Phi(\cdot)$. It will be interesting to find data of more complex input patterns that will require us to increase the capacity in the input encoder $\Phi$. We would like to build upon recent success of understanding information propagation in deep networks using random matrix theory~\cite{pennington2017resurrecting} to further study learning dynamics in these recurrent neural networks. 

\bibliographystyle{plain}
\bibliography{nips_2017}

\begin{thebibliography}{10}

\bibitem{bahdanau2014neural}
Dzmitry Bahdanau, Kyunghyun Cho, and Yoshua Bengio.
\newblock Neural machine translation by jointly learning to align and
  translate.
\newblock {\em arXiv preprint arXiv:1409.0473}, 2014.

\bibitem{chung2014empirical}
Junyoung Chung, Caglar Gulcehre, KyungHyun Cho, and Yoshua Bengio.
\newblock Empirical evaluation of gated recurrent neural networks on sequence
  modeling.
\newblock {\em arXiv preprint arXiv:1412.3555}, 2014.

\bibitem{collins2016capacity}
Jasmine Collins, Jascha Sohl-Dickstein, and David Sussillo.
\newblock Capacity and trainability in recurrent neural networks.
\newblock {\em arXiv preprint arXiv:1611.09913}, 2016.

\bibitem{graves2013speech}
Alex Graves, Abdel-rahman Mohamed, and Geoffrey Hinton.
\newblock Speech recognition with deep recurrent neural networks.
\newblock In {\em Acoustics, speech and signal processing (icassp), 2013 ieee
  international conference on}, pages 6645--6649. IEEE, 2013.

\bibitem{hidasi2015session}
Bal{\'a}zs Hidasi, Alexandros Karatzoglou, Linas Baltrunas, and Domonkos Tikk.
\newblock Session-based recommendations with recurrent neural networks.
\newblock {\em arXiv preprint arXiv:1511.06939}, 2015.

\bibitem{hochreiter1997long}
Sepp Hochreiter and J{\"u}rgen Schmidhuber.
\newblock Long short-term memory.
\newblock {\em Neural computation}, 9(8):1735--1780, 1997.

\bibitem{jozefowicz2016exploring}
Rafal Jozefowicz, Oriol Vinyals, Mike Schuster, Noam Shazeer, and Yonghui Wu.
\newblock Exploring the limits of language modeling.
\newblock {\em arXiv preprint arXiv:1602.02410}, 2016.

\bibitem{karpathy2015visualizing}
Andrej Karpathy, Justin Johnson, and Li~Fei-Fei.
\newblock Visualizing and understanding recurrent networks.
\newblock {\em arXiv preprint arXiv:1506.02078}, 2015.

\bibitem{kiros2015skip}
Ryan Kiros, Yukun Zhu, Ruslan~R Salakhutdinov, Richard Zemel, Raquel Urtasun,
  Antonio Torralba, and Sanja Fidler.
\newblock Skip-thought vectors.
\newblock In {\em Advances in neural information processing systems}, pages
  3294--3302, 2015.

\bibitem{laurent2017chaos}
Thomas Laurent and James~H. von Brecht.
\newblock A recurrent neural network without chaos.
\newblock {\em CoRR}, abs/1612.06212, 2016.

\bibitem{mikolov2010recurrent}
Tomas Mikolov, Martin Karafi{\'a}t, Lukas Burget, Jan Cernock{\`y}, and Sanjeev
  Khudanpur.
\newblock Recurrent neural network based language model.
\newblock In {\em Interspeech}, volume~2, page~3, 2010.

\bibitem{pascanu2013difficulty}
Razvan Pascanu, Tomas Mikolov, and Yoshua Bengio.
\newblock On the difficulty of training recurrent neural networks.
\newblock In {\em International Conference on Machine Learning}, pages
  1310--1318, 2013.

\bibitem{pennington2017resurrecting}
Jeffrey Pennington, Sam Schoenholz, and Surya Ganguli.
\newblock Resurrecting the sigmoid in deep learning through dynamical isometry:
  theory and practice.
\newblock {\em Advances in neural information processing systems}, 2017.

\bibitem{saxe2013exact}
Andrew~M Saxe, James~L McClelland, and Surya Ganguli.
\newblock Exact solutions to the nonlinear dynamics of learning in deep linear
  neural networks.
\newblock {\em arXiv preprint arXiv:1312.6120}, 2013.

\bibitem{wu2017recurrent}
Chao-Yuan Wu, Amr Ahmed, Alex Beutel, Alexander~J Smola, and How Jing.
\newblock Recurrent recommender networks.
\newblock In {\em Proceedings of the Tenth ACM International Conference on Web
  Search and Data Mining}, pages 495--503. ACM, 2017.

\bibitem{xie2017all}
Di~Xie, Jiang Xiong, and Shiliang Pu.
\newblock All you need is beyond a good init: Exploring better solution for
  training extremely deep convolutional neural networks with orthonormality and
  modulation.
\newblock {\em arXiv preprint arXiv:1703.01827}, 2017.

\end{thebibliography}

\end{document}